%% file: main.tex
\documentclass[10pt,twocolumn,letterpaper]{article}

\usepackage{cvpr}              % To produce the CAMERA-READY version
\usepackage{multirow}
\usepackage{amsmath}

\definecolor{cvprblue}{rgb}{0.21,0.49,0.74}
\usepackage[pagebackref,breaklinks,colorlinks,allcolors=cvprblue]{hyperref}

\def\paperID{11197} % *** Enter the Paper ID here
\def\confName{CVPR}
\def\confYear{2025}

\title{Every SAM Drop Counts: Embracing Semantic Priors for \\  Multi-Modality Image Fusion and Beyond}

\author{Guanyao Wu$^{\dag}$, Haoyu Liu$^{\dag}$, Hongming Fu$^{\dag}$, Yichuan Peng$^{\dag}$, Jinyuan Liu$^{\ddag}$, Xin Fan$^{\dag}$, Risheng Liu$^{\dag}$\thanks{Corresponding author.}\\	
	$^\dag$School of Software Technology, Dalian University of Technology, China\\
	$^\ddag$School of Mechanical Engineering, Dalian University of Technology, China\\
	{\tt \small rollingplainko@gmail.com, atlantis918@hotmail.com, \{xin.fan, rsliu\}@dlut.edu.cn}
}
\begin{document}
\maketitle
\input{sec/0_abstract}    
\input{sec/1_intro}

\input{sec/2_RW}

\input{sec/3_method}

\input{sec/4_Exp}
\input{sec/5_Conc}
{
    \small
    \bibliographystyle{ieeenat_fullname}
    \bibliography{main0322}
}


\end{document}

%% file: sec/0_abstract.tex
\begin{abstract}
	
Multi-modality image fusion, particularly infrared and visible, plays a crucial role in integrating diverse modalities to enhance scene understanding. Although early research prioritized visual quality, preserving fine details and adapting to downstream tasks remains challenging. Recent approaches attempt task-specific design but rarely achieve ``The Best of Both Worlds'' due to inconsistent optimization goals. To address these issues, we propose a novel method that leverages the semantic knowledge from the \textbf{S}egment \textbf{A}nything Model (SAM) to \textbf{G}row the quality of fusion results and \textbf{E}nable downstream task adaptability, namely \textbf{SAGE}. Specifically, we design a Semantic Persistent Attention (SPA) Module that efficiently maintains source information via the persistent repository while extracting high-level semantic priors from SAM. More importantly, to eliminate the impractical dependence on SAM during inference, we introduce a bi-level optimization-driven distillation mechanism with triplet losses, which allow the student network to effectively extract knowledge. Extensive experiments show that our method achieves a balance between high-quality visual results and downstream task adaptability while maintaining practical deployment efficiency. The code is available at \url{https://github.com/RollingPlain/SAGE_IVIF}.

\end{abstract}

%% file: sec/1_intro.tex
\section{Introduction}
\label{sec:intro}

\begin{figure}[t]
	\centering
	\includegraphics[width=1\linewidth]{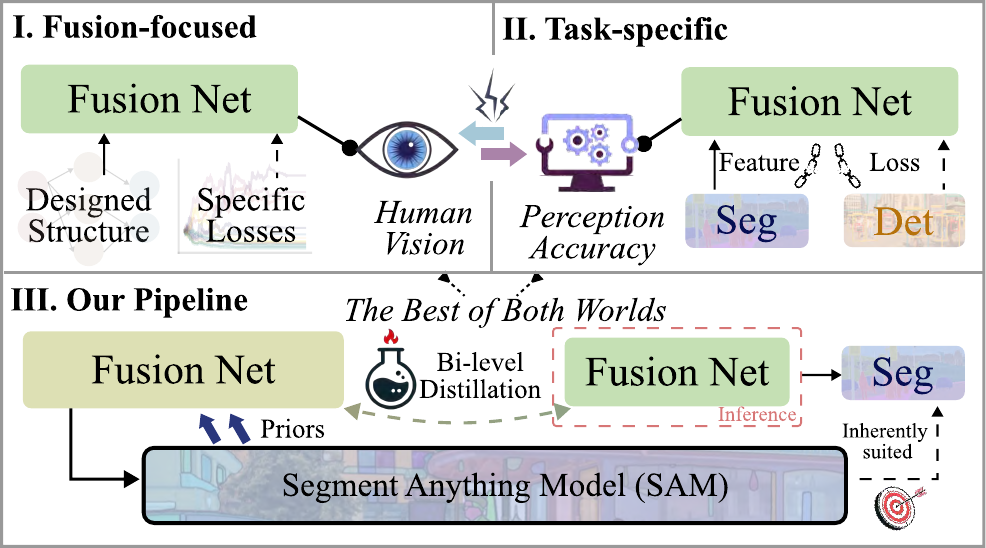 }
	
	\caption{Differences between the proposed method and existing mainstream comparative approaches: (a) Traditional and early DL-based methods focus on the fusion visual effect. (b) Task-specific methods (e.g., TarDAL~\cite{liu2022target} \& SegMiF~\cite{liu2023multi}) introduce task loss and features that lead to inconsistent optimization goals, causing a conflict between visual and task accuracy. (c) Our pipeline first leverages semantic priors from SAM within a large network and then distills the knowledge into a smaller sub-network achieving practical inference feasibility while ensuring ``the best of both worlds" through SAM's inherent adaptability to these tasks.}
	\label{fig:first}
	\vspace{-0.5cm}
\end{figure}

In light of recent advancements in sensing technologies, multi-modality imaging~\cite{liu2024infrared} has gained significant attention, with applications in robotics~\cite{zhang2021image}, remote sensing~\cite{xu2021deep}, and autonomous driving~\cite{bijelic2020seeing}.
Integrating infrared and visible sensors provides substantial advantages for intelligent processing~\cite{kaur2021image}, although both exhibit inherent limitations~\cite{ma2019infrared, liu2019convergence}. Infrared images are robust against smoke, obstruction, and low light, while visible images excel in resolution, contrast, and texture detail. These characteristics are highly complementary, motivating their fusion to generate a comprehensive fused image that balances visual quality and downstream task requirements. Consequently, achieving this fusion efficiently remains a pressing challenge~\cite{liu2024infrared}.

Traditional Infrared and Visible Image Fusion (IVIF) methods, based on information theory~\cite{zhang2020vifb, liu2020bilevel}, aim to retain as much source image information as possible but often struggle with optimizing fused image quality, especially in managing redundancy and specific scenarios. Early deep learning-based methods~\cite{xu2022u2fusion, li2023lrrnet, liu2023coconet, zhao2023ddfm, zhao2023cddfuse, ma2022toward, liu2022twin} focused on visualization but faced issues like edge blurring and artifacts, failing to meet downstream perception needs. Current methods~\cite{liu2022target, liu2023multi, tang2022image, liu2023bi} focus on specific tasks, coupling fusion with downstream processing, resulting in conflicting optimization objectives and challenges in balancing visual quality with task adaptability.

\begin{figure}[t]
	\centering
	\includegraphics[width=1\linewidth]{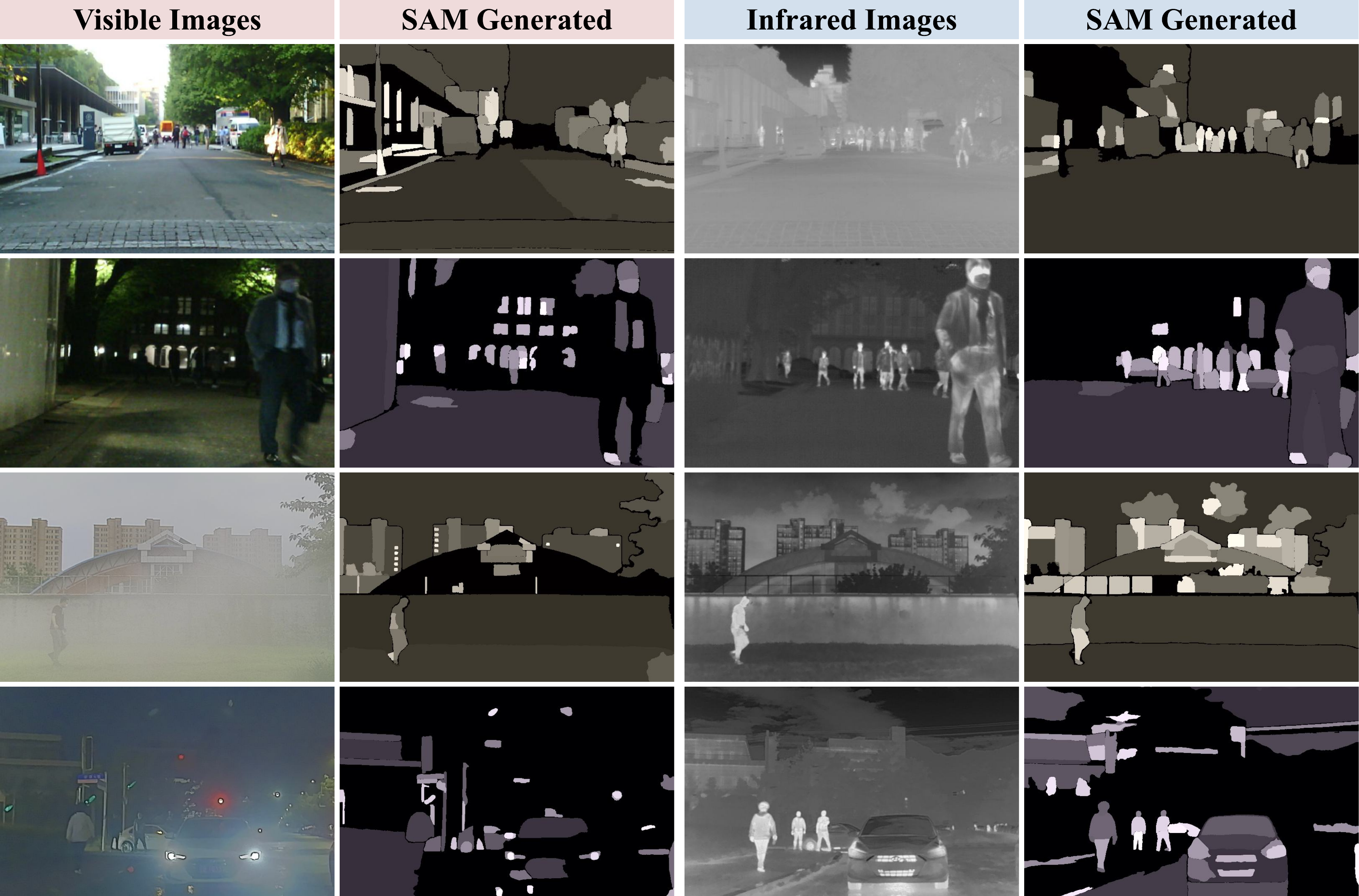}
	
	\caption{Demonstration of SAM's robustness in MFNet normal scenes (top) and under challenging FMB conditions (bottom).}
	\label{fig:robustness}
	\vspace{-0.4cm}
\end{figure}

Recent advancements in large-scale visual models have significantly improved various visual analysis tasks~\cite{wang2023large}. Among these, the Segment Anything Model~\cite{kirillov2023segment} stands out due to its exceptional and robust capability to provide rich semantic information, making it well-suited for IVIF, as illustrated in Figure~\ref{fig:robustness}. Also, its segmentation capabilities align naturally with the requirements of downstream tasks in IVIF field. However, current methods that integrate SAM for low-level tasks typically require full SAM during inference, resulting in excessive practical infeasibility.

To address these issues, we propose the fusion method SAGE that fully integrates and distills semantic priors derived from SAM. As illustrated in Figure~\ref{fig:first} III, one of our aim is to harness the advantages of SAM's semantic priors while reducing computational burden. The proposed method consists of two key components: Semantic Persistent Attention module and a bi-level distillation scheme. The former focuses on using a persistent repository throughout the process to retain source image information, while heuristically enabling the network to integrate semantic patches predicted by SAM.
Furthermore, the distillation scheme is designed to efficiently exclude SAM from the inference process, thereby reducing computational complexity. By enhancing distilled information across different aspects, the triplet losses in the scheme significantly boost the performance of the fusion model. Consequently, efficient fusion is achieved using only the distilled model, delivering high-quality visual results and precise task performance without the direct involvement of SAM. In summary, our contributions can be summarized as follows:

\begin{itemize}
	
	\item A novel fusion framework is proposed to leverage semantic priors from the SAM, effectively balancing visual quality and downstream task adaptability..
	
	\item The designed Semantic Persistent Attention (SPA) module utilizes a persistent repository to efficiently retain source information and simultaneously extract high-level semantic representation.
	
	\item We develop a bi-level optimization distillation scheme that transfers the information processed by SPA into the sub-network, effectively decoupling the fusion process from SAM during inference.
	
	\item Extensive experiments on multiple datasets validate the superior effectiveness of our proposed method.
\end{itemize}

%-------------------------------------------------------------------------

%% file: sec/2_RW.tex
\section{Related Works}
\label{sec:related}

\textbf{Learning based IVIF Methods.}~In recent years, deep learning based multi-modality image fusion approaches~\cite{xu2022u2fusion, li2023lrrnet, liu2023coconet, zhao2023ddfm, zhao2023cddfuse, zhou2022unified} achieved significant progress, primarily focusing on improving fusion quality, such as enhancing image details, maintaining structural consistency, and reducing noise. These approaches, including innovations in adversarial learning, Transformer models and diffusion network, have provided new insights. However, most of these methods have not addressed how the fused images can directly contribute to downstream tasks such as object detection, semantic segmentation, and other perception-related applications. With the growing demand for more integrated solutions, this gap has sparked increasing research interest in combining fusion with various downstream tasks.

Early works began to bridge this gap, with TarDAL~\cite{liu2022target} pioneering the use of a dual-discriminator network for object detection optimization, which simultaneously considers modality differences and detection tasks during fusion. As the field has progressed, several methods~\cite{sun2022detfusion, zhao2023metafusion} have also focused on optimizing fusion for object detection. For semantic segmentation~\cite{zhang2024mrfs},  SeAFusion~\cite{tang2022image} and SegMiF~\cite{liu2023multi} have made early significant improvements, with the former enhancing fusion via task loss and the latter leveraging task-specific features. More recently, some approaches~\cite{liuelegance, liu2023bi, tang2022superfusion, liu2024task, jiang2024multispectral, wang2022unsupervised} have integrated multi-task learning to jointly optimize detection and segmentation, improving performance on both tasks.

\begin{figure*}[!htb]
	\centering
	\setlength{\tabcolsep}{1pt} 
	
	\includegraphics[width=0.99\textwidth]{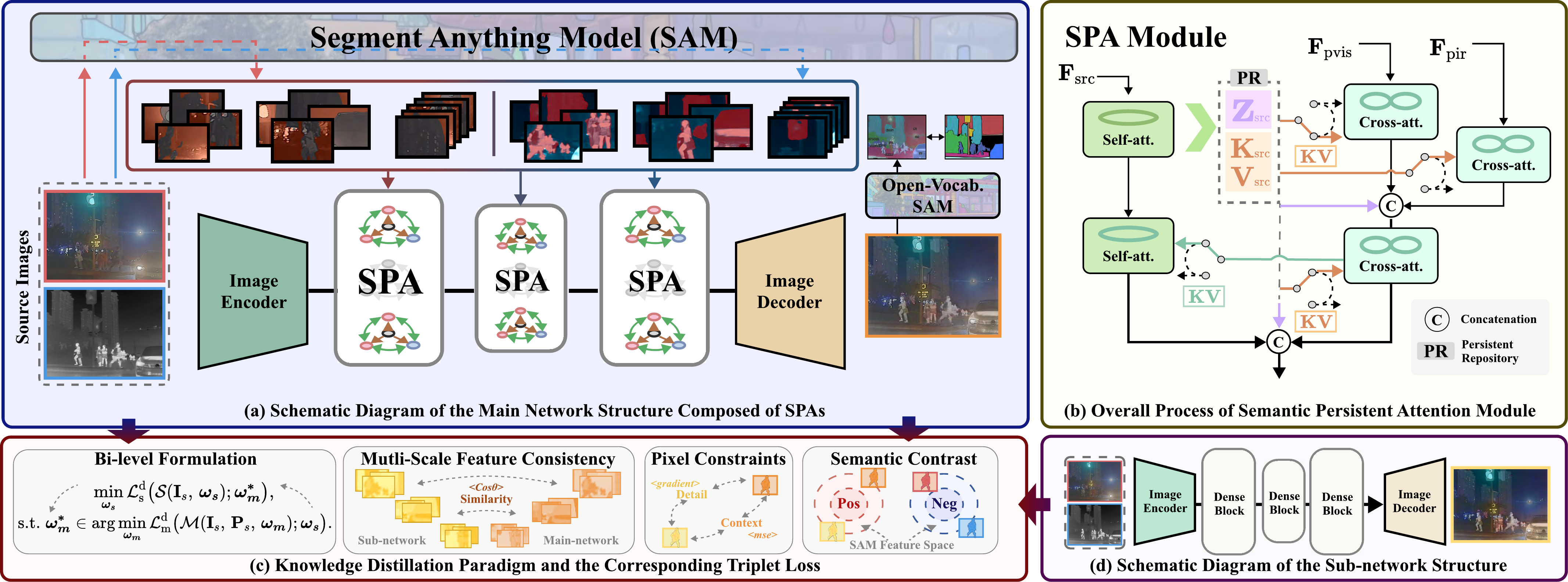}
	
	\caption{A overall workflow of our proposed method. (a) shows the flow structure of the main network, where the SPA module processes patches with semantic priors generated by SAM. (b) illustrates the detailed structure of the SPA module, where PR plays a key role in preserving the source and integrating the semantic information. (c) displays our distillation scheme formulation, with visualizations of the different components of the triplet loss. (d) provides a simple diagram of the sub-network, which is composed of stacked dense blocks.}
	\label{fig:workflow}
\end{figure*}

\noindent \textbf{SAM \& Its Application.}~
The Segment Anything Model (SAM)~\cite{kirillov2023segment} is a large-scale pre-trained foundation model that excels in zero-shot generalization, enabling it to handle a wide range of segmentation and detection tasks without the need for task-specific training. SAM demonstrates impressive performance in low-level vision applications, such as image deblurring~\cite{li2024sam}, dehazing~\cite{jin2023let}, super-resolution~\cite{wang2024sam}, and low-light image enhancement~\cite{li2024low}. Moreover, SAM’s capabilities extend to high-level tasks like object detection~\cite{ren2024grounded}, which further strengthens its versatility. This ability to handle both low-level and high-level tasks makes SAM inherently well-suited for the needs of the IVIF field, where fusion based segmentation and detection tasks are critical. However, its impressive performance is hindered by high computational complexity, posing challenges for practical deployment. Addressing these computational burdens, while fully leveraging the semantic priors of SAM, is a primary motivation for our work.

%% file: sec/3_method.tex
\section{The Proposed Method}
\subsection{Problem Formulation}

\begin{figure*}[!htb]
	\centering
	\setlength{\tabcolsep}{1pt} 
	
	\includegraphics[width=0.99\textwidth]{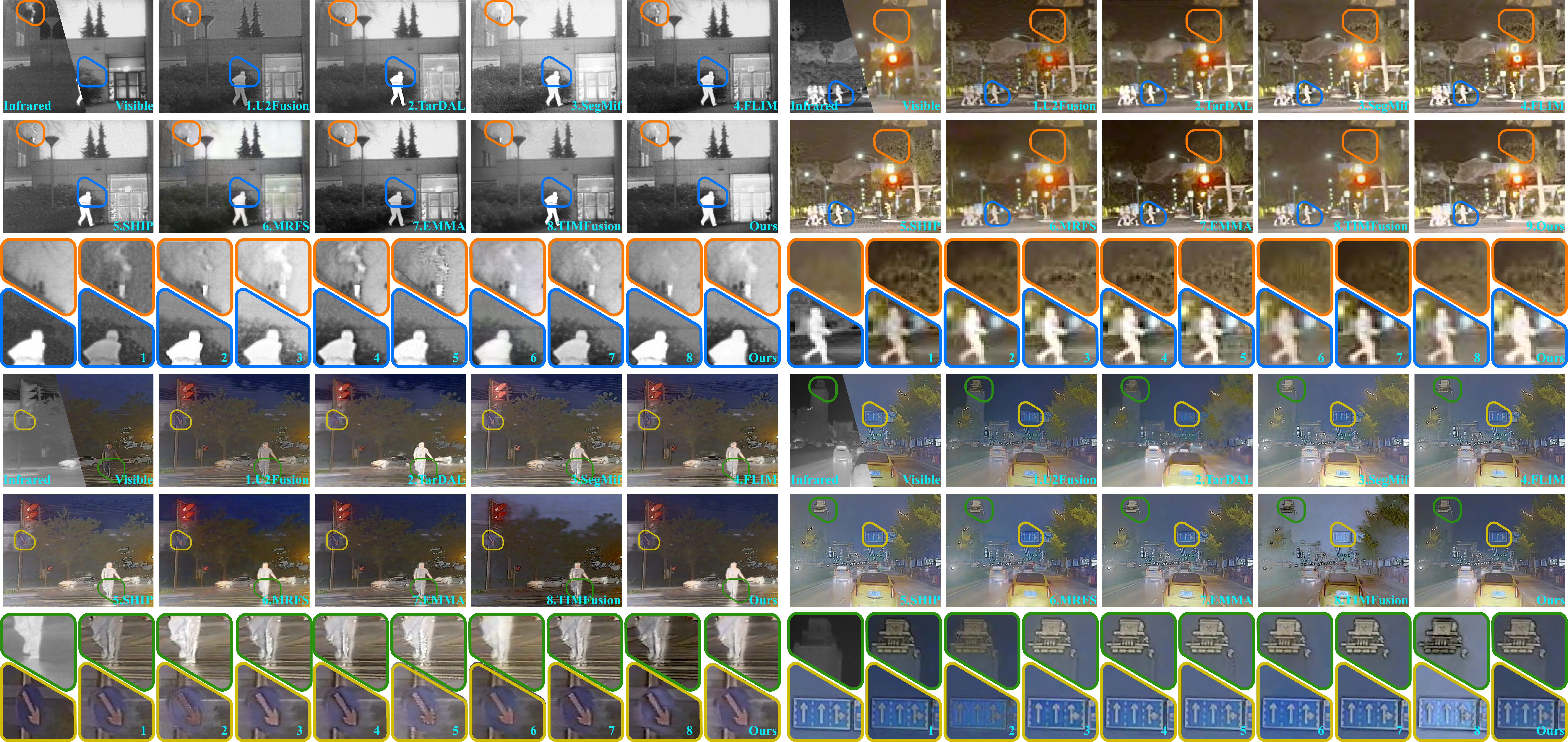}
	
	\caption{Qualitative demonstrations of SOTA approaches across commonly used datasets, including TNO, RoadScene, M$^3$FD and FMB.}
	\label{fig:fusion_main}
\end{figure*}

The overall workflow of our method is illustrated in Figure~\ref{fig:workflow}. Our aim is to fully exploit the semantic priors of SAM during the inference stage to enhance cross-modal fusion quality. However, directly deploying a large-scale SAM model often leads to prohibitive computational overhead.
To mitigate this, we adopt a knowledge distillation strategy, transferring the information encoded by the SAM-driven main network to a lightweight sub-network, thereby significantly reducing inference costs while maintaining high-quality fusion.
Nevertheless, the substantial capacity gap between the SAM-enhanced main network and the compact sub-network frequently results in incomplete semantic transfer or structural inconsistencies, which hinder ideal cross-modal fusion performance.
To address this issue, we propose a bi-level optimization framework that jointly optimizes both networks as a unified system, aiming to bridge the distillation gap and maintain consistent fusion guided by SAM's semantic priors.

Let $\mathbf{I}_\mathrm{s}$ denote the source visible and infrared input images, \emph{i.e.}, $\mathbf{I}_\mathrm{s} = \{\mathbf{I}_\mathrm{vis}, \mathbf{I}_\mathrm{ir}\}$. Their corresponding semantic-prior patches generated under the guidance of SAM are denoted as $\mathbf{P}_\mathrm{s} = \{\mathbf{P}_\mathrm{vis}, \mathbf{P}_\mathrm{ir}\}$.
Denote the main network by $\mathcal{M}$ with parameters $\boldsymbol{\omega_\mathrm{m}}$, and the sub-network by $\mathcal{S}$ with parameters $\boldsymbol{\omega_\mathrm{s}}$. We define the optimization process as:
\vspace{-0.2cm}
\begin{equation}
	\label{eq:low}
	\begin{gathered}
		\min_{\boldsymbol{\omega_s}} \mathcal{L}^\mathrm{d}_\mathrm{s} \big(\mathcal{S}(\mathbf{I}_s,\, \boldsymbol{\omega_s}); \boldsymbol{\omega_m^*}\big),\\
		\text{s.t.} \quad \boldsymbol{\omega_m^*} \in \arg\min_{\boldsymbol{\omega_m}} \mathcal{L}^\mathrm{d}_\mathrm{m} \big(\mathcal{M}(\mathbf{I}_s,\, \mathbf{P}_s,\, \boldsymbol{\omega_m}); \boldsymbol{\omega_s}\big),
	\end{gathered}
\end{equation}
\vspace{-0.3cm}

\noindent where $\mathcal{L}^\mathrm{d}_\mathrm{m}$ is the distillation loss that ensures high-quality fusion with meaningful semantic cues in the main network, and $\mathcal{L}^\mathrm{d}_\mathrm{s}$ is the distillation loss guiding the sub-network to effectively mimic the main network's behavior. The fused reference image is given by: $\mathbf{I}_\mathrm{ref} = \mathcal{M}\big(\{\mathbf{I}_\mathrm{vis}, \mathbf{I}_\mathrm{ir}\}, \{\mathbf{P}_\mathrm{vis}, \mathbf{P}_\mathrm{ir}\}, \boldsymbol{\omega_m}\big)$, while the sub-network produces a fused image $\mathbf{I}_\mathrm{f} = \mathcal{S}\big(\mathbf{I}_\mathrm{vis}, \mathbf{I}_\mathrm{ir}, \boldsymbol{\omega_s}\big)$.
Notably, both distillation objectives rely on outputs from the counterpart network, establishing a bidirectional dependency, which is a key feature of our distillation formulation.

Within the overall framework, the main network integrates SAM-derived semantic priors into the fusion process, while the sub-network is optimized to align with its outputs under the bi-level formulation. The following sections elaborate on two key components that support this process: the Semantic Persistent Attention (SPA) module~\ref{sec:3.2}, which integrates SAM-derived priors with source features, and the triplet-based loss scheme~\ref{sec:3.3}, which facilitates structured knowledge transfer across networks.

\subsection{Semantic Persistent Attention Module}
\label{sec:3.2}
To fully exploit the semantic information provided by SAM, we propose the Semantic Persistent Attention (SPA) module, as illustrated in the flowchart in Figure~\ref{fig:workflow} (b).

The key core of the SPA module is a Persistent Repository (PR), which serves as a static memory to store and maintain crucial contextual information for the fusion process. Specifically, PR stores the latent representation (\(\textbf{Z}\)) of the source features \(\textbf{F}_\text{src}\) and the corresponding key-value pairs (\(\textbf{K}_\text{src}, \textbf{V}_\text{src}\)), which provide consistent contextual support during the cross-attention operation. The core insight behind this design is that PR acts as a stable, modality-specific information source, guiding the cross-attention mechanism to fuse the semantic patches with rich contextual details from the original source.

The visible and infrared semantic patches (\(\textbf{P}_\text{vis}\) and \(\textbf{P}_\text{ir}\)) extracted from the SAM are encoded into features (\(\textbf{F}_\text{pvis}\) and \(\textbf{F}_\text{pir}\)). These encoded semantic features represent limited portions of the scene and are processed through the cross-attention mechanism. By utilizing the key-value pairs (\(\textbf{K}_\text{src}, \textbf{V}_\text{src}\)) stored in PR, the cross-attention mechanism enriches the semantic queries (\(\textbf{Q}_\text{pvis}\) and \(\textbf{Q}_\text{pir}\)) with the full scene context from the source image, thereby addressing the inherent incomplete scene coverage in the patches. PR ensures the fusion process stays grounded in a stable, consistent context, enabling patches to be effectively enriched with modality-specific information. This stability maintains semantic coherence while allowing flexible feature refinement via attention mechanisms.

In summary, the SPA module uses PR to guide the cross-attention mechanism, ensuring that the semantic patches from SAM are enriched with consistent and detailed modality-specific information. This approach allows the fusion of infrared and visible image modalities by combining both the high-level semantic understanding from SAM and the detailed information from the source features. The final output, \(\textbf{F}_\text{SPA}\), represents a semantically enriched, structurally consistent feature set, rich in semantic priors of SAM, and ready for further processing.

The main network heavily utilizes the SPA module, aiming to fully exploit the complex semantic priors provided by SAM, focus on capturing and retaining detailed semantic knowledge, thereby facilitating a powerful representation that can later be distilled into a more efficient sub-network.

\begin{figure*}[!htb]
	\centering
	\setlength{\tabcolsep}{1pt} 
	
	\includegraphics[width=0.99\textwidth]{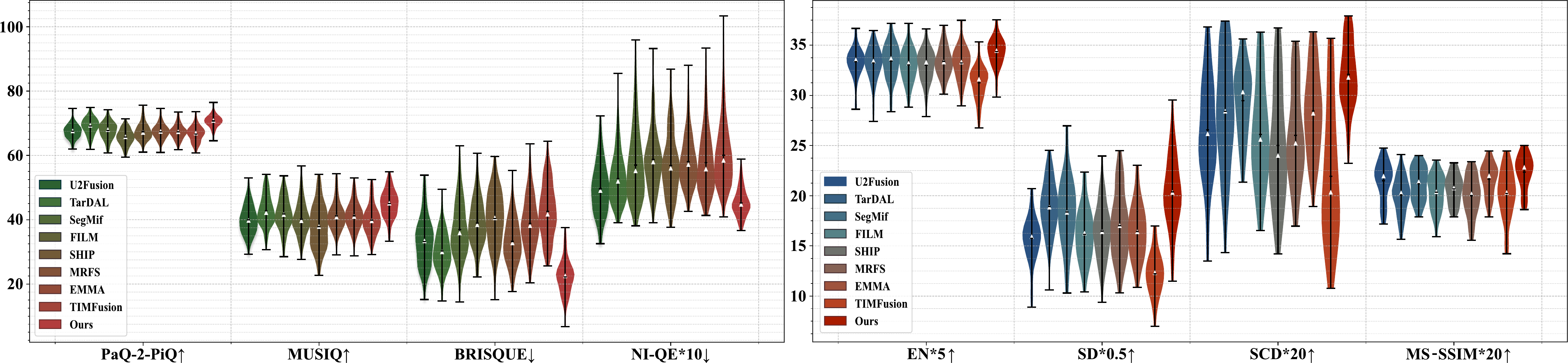}
	
	\caption{Quantitative comparison of fusion performance with other SOTA methods on M$^3$FD and FMB benchmarks. The violin plots illustrate the distribution of metrics, in which the the black lines and white triangles indicate the medium and mean values.}
	\label{fig:fusion_metric}
\end{figure*}

\subsection{Triplet Loss in Distillation Scheme}
\label{sec:3.3}
In this section, we introduce a triplet loss-driven distillation scheme that facilitates semantic knowledge transfer from the main network (\(\mathcal{M}\)) to the sub-network (\(\mathcal{S}\)) under a bi-level optimization framework. Specifically, we adopt a DARTS-style~\cite{liu2018darts} training protocol, where the main network (teacher) and the sub-network (student) are updated in small alternating steps, approximating the inner-outer structure of bilevel optimization. 
This allows gradients to flow in a bidirectional manner: 
the sub-network learns from the distillation signals provided by the main network, 
while the main network also adapts its parameters based on the sub-network’s performance and a segmentation objective. 
Consequently, the two networks reach a cohesive transfer with mutual compromise.
 
Our distillation scheme is driven by three distinct losses, each focusing on a different aspect of the fused images. 
Let \(\mathbf{I}_\text{ref}\) be the fused reference image produced by the main network, and \(\mathbf{I}_\text{f}\) be the fused output from the sub-network.
We first introduce a feature alignment term:

\begin{equation}
	\mathcal{L}_{\text{fea}} = \sum_{m=1}^{M} \left( 1 - \frac{\textbf{F}_\text{Den}^m \cdot \textbf{F}_\text{SPA}^m}{\|\textbf{F}_\text{Den}^m\|_2 \cdot \|\textbf{F}_\text{SPA}^m\|_2} \right),
\end{equation}
where \(\textbf{F}_\text{Den}^m\) and \(\textbf{F}_\text{SPA}^m\) are the feature maps of the dense and SPA blocks at the same scale \(m\), and \(\|\cdot\|_2\) is the $\ell_2$-norm.

The second loss component operates at the context level, which consists of two parts: a gradient part to preserve structural consistency and an MSE loss for intensity consistency. They are defined as:

\begin{equation}
\mathcal{L}_\text{grad} = \left\|\nabla \mathbf{I}_\text{ref} - \nabla \mathbf{I}_\text{fus}\right\|_1, \quad \mathcal{L}_\text{MSE} = \left\| \mathbf{I}_\text{ref} - \mathbf{I}_\text{fus}\right\|_2,
\end{equation}
where \(\nabla\) denotes the Sobel operator and \(\|\cdot\|_1\) is the $\ell_1$
-norm. So the cont loss can be calculated as $\mathcal{L}_\text{context} = \mathcal{L}_\text{grad} + \mathcal{L}_\text{MSE}$. In practice, we apply these terms not only between \(\mathbf{I}_\text{ref}\) and \(\mathbf{I}_\text{fus}\), but also between each fused output and the original source images to ensure the reconstruction fidelity. This prevents both networks from drifting away from the source image during the distillation process.

The third loss component, contrastive semantic loss aims to ensure that the feature space of the fused image remains close to that of the reference image, while being distinct from the individual visible and infrared images archiving efficient distillation. We utilize the semantic feature space by encoder of SAM (defined as $\mathcal{S}_\text{E}$) to construct positive and negative pairs through the element-wise multiplication of the binary masks ($\mathbf{M}_{\text{vis}}$ and $\mathbf{M}_{\text{ir}}$) from $\textbf{P}_\text{vis}$ and $\textbf{P}_\text{ir}$. The contrastive semantic loss on each modality is defined as:

\begin{equation}
\mathcal{L}_\text{cs}^\text{ir} = \sum_{l=1}^{L}  \frac{\left\| \mathcal{S}_\text{E}(\mathbf{I}_{\text{fus}} \odot \mathbf{M}_{\text{ir}}) - \mathcal{S}_\text{E}(\mathbf{I}_{\text{ref}} \odot \mathbf{M}_{\text{ir}}) \right\|_2}{\left\| \mathcal{S}_\text{E}(\mathbf{I}_{\text{x}} \odot \mathbf{M}_{\text{ir}}) - \mathcal{S}_\text{E}(\mathbf{I}_{\text{ir}} \odot \mathbf{M}_{\text{ir}}) \right\|_2},
\end{equation}

\begin{equation}
\mathcal{L}_\text{cs}^\text{vis} = \sum_{l=1}^{L} \frac{\left\| \mathcal{S}_\text{E}(\mathbf{I}_{\text{fus}} \odot \mathbf{M}_{\text{vis}}) - \mathcal{S}_\text{E}(\mathbf{I}_{\text{ref}} \odot \mathbf{M}_{\text{vis}}) \right\|_2}{\left\| \mathcal{S}_\text{E}(\mathbf{I}_{\text{x}} \odot \mathbf{M}_{\text{vis}}) - \mathcal{S}_\text{E}(\mathbf{I}_{\text{vis}} \odot \mathbf{M}_{\text{vis}}) \right\|_2},
\end{equation}
where $l$ indexes the layers of $\mathcal{S}_\text{E}$, and x $\in \{\text{ref}, \text{fus}\}$. So the total $\mathcal{L}_\text{cs} = \mathcal{L}_\text{cs}^\text{ir} + \mathcal{L}_\text{cs}^\text{vis}$ can also be computed bidirectionally in the bi-level optimization scheme between two networks, ensuring that the feature spaces of both the \(\mathbf{I}_\text{ref}\) and the  \(\mathbf{I}_\text{f}\) are easily aligned.

The total distillation loss for the sub-network, denoted as $\mathcal{L}^\text{d}_\text{s}$ in Eq.~\eqref{eq:low}, is the sum of the feature loss, context loss, and contrastive semantic loss: $\mathcal{L}^\text{d}_\text{s} = \mathcal{L}_{\text{fea}} + \mathcal{L}_{\text{cont}} + \mathcal{L}_{\text{cs}}$.

For the main network, an additional segmentation cross-entropy loss \(\mathcal{L}_\text{seg}\) is incorporated. This loss is computed between the segmentation predictions generated by an open-vocabulary model \(\mathcal{S}_\text{O}\)~\cite{zou2023generalized, zou2023segment} and the ground truth segmentation labels. Instead of affecting the sub-network, this loss aim to prevent potential optimization conflicts. The segmentation loss \(\mathcal{L}_\text{seg}\) is defined as:

\begin{equation}
	\mathcal{L}_\text{seg} = - \sum_{c} \left[ \mathbf{Y}_\text{label}^{c} \log(\mathbf{Y}_\text{pred}^{c}) \right],
\end{equation}
where \(\mathbf{Y}_\text{label}\) represents the label segmentation map, \(\mathbf{Y}_\text{pred}\) is the predicted segmentation map, and \(c\) is the class index. For now, the total distillation loss for the main network can be calculated as: $\mathcal{L}^\text{d}_\text{m} = \mathcal{L}^\text{d}_\text{s} + \mathcal{L}_\text{seg}$.

%% file: sec/4_Exp.tex
\section{Experiments}

\begin{table*}
	\centering
	\small
	\renewcommand\arraystretch{1} 
	\setlength{\tabcolsep}{1.2mm}
	\begin{tabular}{|l|l|ccccccc|c|ccccccc|c|}  
		\toprule
		\multicolumn{2}{|c|}{Framework } & \multicolumn{8}{c|}{Segformer - B3~\cite{xie2021segformer}} & \multicolumn{8}{c|}{X - Decoder~\cite{zou2023generalized}}  \\  
		\bottomrule
		\toprule
		Method & Venue & Road & Bldg. & Sign & Veg. & Bus & Per. & Car & mIoU & Road & Bldg. & Sign & Veg. & Bus & Per. & Car & mIoU \\  
		\hline
		Visible &- & 84.8 & 81.0 & 64.2 & 84.7 & 46.8 & 61.1 & 77.1 & 49.1 & 79.2 & 67.8 & 18.2 & 73.2 & 52.0 & 56.9 & 82.0 & 50.1 \\
		
		Infrared &-  & 83.7 & 77.6 & 55.9 & 80.9 & 38.1 & 57.1 & 74.2 & 42.7 & 76.6 & 55.6 & 12.9 & 38.2 & 72.7 & 63.2 & 78.7 & 42.9 \\
		
		\bottomrule
		\toprule
		
		DDFM & ICCV'23 & 86.2 & \textcolor{blue}{\textbf{83.0}} & 69.5 & 84.8 & 62.5 & 61.0 & 80.8 & \textcolor{blue}{\textbf{58.2}} & 76.1 & 65.9 & 10.1 & 73.3 & 72.0 &  \textcolor{red}{\textbf{66.6 }}& 82.5 & 49.3 \\
		
		U2Fusion & TPAMI'22 & 86.7 & 82.4 & 68.5 & 84.6 & 50.3 & 61.8 & 80.4 & 57.0 & \textcolor{blue}{\textbf{79.7}}  & \textcolor{red}{\textbf{70.6}} & 22.2 & 72.7 & 75.7 & 65.5 & \textcolor{red}{\textbf{83.3}} & 50.4 \\
		
		TarDAL & CVPR'22 & 85.8 & 81.8 & 68.4 & 84.7 & 49.5 & 61.3 & 79.5 & 55.7 & 74.1 & 63.1 & 23.2 & 67.5 & 74.8 & \textcolor{blue}{\textbf{66.4 }} & 81.1 & 48.7 \\
		
		SegMiF & ICCV'23 & 86.3 & 82.2 & 68.0 & 85.0 & \textcolor{blue}{\textbf{63.7}}  & 60.7 & 80.0 & 57.3  & 79.6 & 68.6 & 14.0 & 73.8 & 74.3 & 65.8 &\textcolor{blue}{\textbf{83.2}} & \textcolor{blue}{\textbf{50.8}}  \\
		
		FILM &ICML'24 & 86.9 & 81.9 & 69.1 & 84.0 & 46.1 & 60.6 & 79.4 & 55.7 & 79.2 & 70.3 & 17.9 & 73.7 & \textcolor{red}{\textbf{79.0}}  & 64.0 & 82.0 & 50.6 \\
		
		SHIP &CVPR'24 & 84.9 & 82.1 & 68.3 & 83.3 & 61.4 & 60.9 & 80.4 & 57.2 & 76.4 & 66.8 & \textcolor{red}{\textbf{24.7}} & 71.2 & 76.3 & 64.6 & 80.9 & 50.4 \\
		
		MRFS &CVPR'24 & \textcolor{blue}{\textbf{88.2}} & 82.8 & \textcolor{blue}{\textbf{71.0}} & \textcolor{blue}{\textbf{86.0}} & 61.5 & \textcolor{blue}{\textbf{63.6}} & \textcolor{blue}{\textbf{81.1}} & 56.5 & 73.8 & 65.0 & \textcolor{blue}{\textbf{23.7}} & 72.3 & 77.2 & 66.2 & 82.6 & 49.6 \\
		
		EMMA & CVPR'24 & 85.2 & 81.2 & 66.8 & 83.3 & 59.0 & 58.9 & 79.9 & 55.8 & 78.6 & 68.4 & 15.0 & \textcolor{blue}{\textbf{74.3}} & 77.1 & 64.5 & 82.2 & 50.7 \\
		
		TIMFusion & TPAMI'24 & 85.1 & 80.0 & 66.4 & 82.8 & 50.8 & 59.0 & 79.7 & 55.5 & 62.0 & 63.0 & 12.8 & 67.6 & 75.8 & 65.8 & 79.6 & 46.1 \\
		
		SAGE & Proposed & \textcolor{red}{\textbf{89.0}} & \textcolor{red}{\textbf{83.9}} & \textcolor{red}{\textbf{71.4}} & \textcolor{red}{\textbf{86.3}} & \textcolor{red}{\textbf{66.7}} & \textcolor{red}{\textbf{64.1}} & \textcolor{red}{\textbf{81.7}} & \textcolor{red}{\textbf{61.2}} & \textcolor{red}{\textbf{81.0}} & \textcolor{blue}{\textbf{70.4}} &17.0 & \textcolor{red}{\textbf{75.2}} & \textcolor{blue}{\textbf{77.7}} &64.0 & 82.8 & \textcolor{red}{\textbf{51.1}}  \\
		
		\bottomrule
	\end{tabular}
	
	\caption{Quantitative semantic segmentation results of SOTA approaches on the FMB dataset.}
	\label{tab:123}
\end{table*}

\begin{figure*}[!htb]
	\centering
	\setlength{\tabcolsep}{1pt} 
	
	\includegraphics[width=0.99\textwidth, height=0.21\textheight]{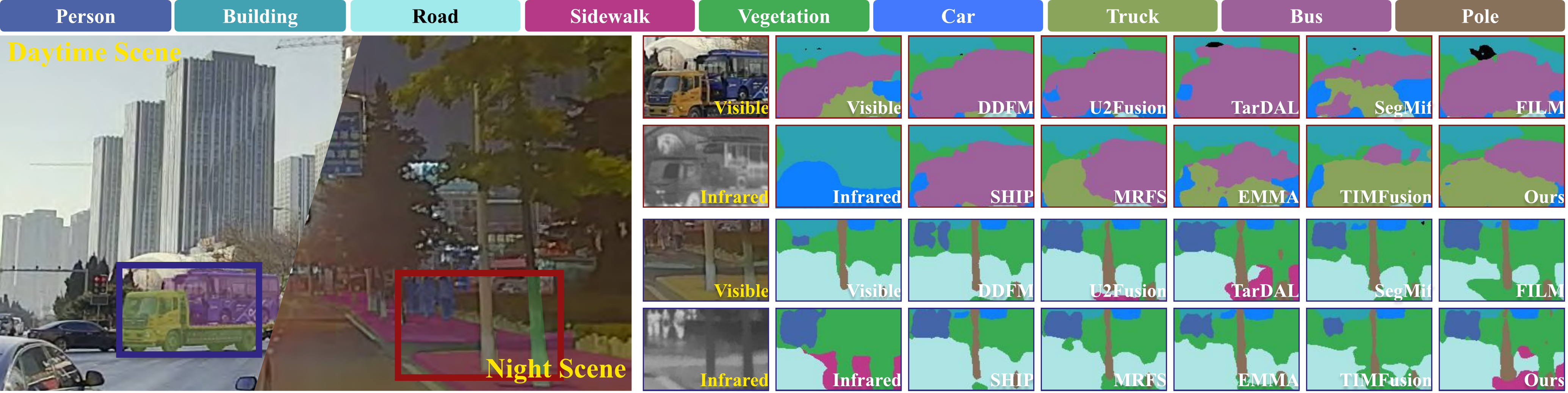}
	
	\caption{Qualitative segmentation results of SOTA approaches on the FMB dataset.}
	\label{fig:seg_main_fmb}
	\vspace{-0.2cm}
\end{figure*}

\subsection{Settings and Details}

Five representative datasets, namely TNO~\cite{toet2017tno}, RoadScene~\cite{xu2022u2fusion}, MFNet~\cite{ha2017mfnet}, FMB~\cite{liu2023multi}, and M$^3$FD~\cite{liu2022target}, are utilized for both training and evaluation. For segmentation, SegFormer~\cite{xie2021segformer} (B3 variant) is adopted as the backbone, and the models are trained for 100 epochs. The training and testing splits are strictly in accordance with the official dataset guidelines. Adam is adopted for training, with initial learning rates of $5 \times 10^{-4}$ and $2 \times 10^{-3}$ for the main and sub-networks, respectively. Cosine annealing decays both to $1 \times 10^{-5}$. Prior to distillation, the models undergo a pre-training phase, followed by 5 epochs of distillation. The batch size is set to 4, and the images are randomly cropped and resized to \(192 \times 256\) during training. The entire framework is implemented in PyTorch and executed on two NVIDIA GeForce RTX 4090 GPUs.

\subsection{Results of Multi-modality Image Fusion}
We demonstrate our fusion quality by visualization and quantitative analysis with 9 other SOTA methods in recent years, including
DDFM~\cite{zhao2023ddfm}, U2Fusion~\cite{xu2022u2fusion}, TarDAL~\cite{liu2022target}, SegMiF~\cite{liu2023multi}, FILM~\cite{zhao2024image}, SHIP~\cite{zheng2024probing}, MRFS~\cite{zhang2024mrfs}, EMMA~\cite{zhao2024equivariant} and TIMFusion~\cite{liu2024task}.

\noindent \textbf{Qualitative Comparisons.} Figure~\ref{fig:fusion_main} presents a visual comparison of various methods. Overall, the proposed method has two main advantages. First, it effectively preserves the multi-modality information of the original images. In the TNO surveillance scene, the vegetation details from the visible image and the smoke from the chimney in the infrared image are both well retained (the image set in the top-left corner). In RoadScene, our method also achieves optimal leaf recovery. On the other hand, our approach demonstrates strong robustness to interference, as it accurately reconstructs the reflective pedestrian crossing lines on the ground and the distant building outlines in the dense fog at night (the green box in the second row). The integration of SAM enhances our method, giving it the ability to surpass other methods and achieve superior results.

\noindent \textbf{Quantitative Comparisons.}~We also compare our numerical results with other competitors in Figure~\ref{fig:fusion_metric}, based on 100 image pairs randomly selected from the FMB and M$^3$FD datasets. We employ four fusion evaluation metrics~\cite{ma2019infrared}, including the Entropy (EN), Standard Deviation (SD), Sum of Correlation Differences (SCD), and Multi-Scale Structural Similarity (MS-SSIM). Additionally, we introduce four no-reference image quality assessments~\cite{ying2020patches} related to image quality: BRISQUE, NIQE, MUSIQ, and PaQ-2-PiQ.

In terms of these metrics, our method demonstrates consistent superiority. High SCD and MS-SSIM values indicate that our method effectively preserves a significant amount of information from the source images, while EN reflects our rich detail and texture. The spatial, multi-scale, and patch evaluations of the NR-IQA confirm that our fusion approach has a better pixel distribution and aligns more closely with the human visual system perception.

\begin{table}
	\centering
	\small
	\renewcommand\arraystretch{0.9} 
	\setlength{\tabcolsep}{1mm}
	\begin{tabular}{|l|ccccccc|c|}  
		\toprule
		\multicolumn{1}{|c|}{Framework } & \multicolumn{8}{c|}{Segformer - B3~\cite{xie2021segformer}} \\  
		\bottomrule
		\toprule
		Method  & Car & Per. & Bike & Curve & C.S. & C.C. & Bump & mIoU \\   
		\hline
		Visible  & 83.3 & 54.7 & 59.6 & 7.46 & 31.0 & 37.4 & 30.8 & 44.6  \\
		
		Infrared  & 79.0 & 63.8 & 51.4 & 7.10 & 25.2 & 34.1 & 41.8 & 44.4 \\
		
		\bottomrule
		\toprule
		
		DDFM  & 86.7 & 62.2 & 63.4 & 30.8 & \textcolor{blue}{\textbf{35.5 }}& 45.7 & 41.1 & 51.7 \\
		
		U2Fusion  & \textcolor{blue}{\textbf{87.3}}  & 65.9 & 60.8 & 33.4 & \textcolor{red}{\textbf{36.0}} & 46.1 & 40.9 & 52.0  \\
		
		TarDAL  & 86.4 & \textcolor{blue}{\textbf{67.1}} & 62.4 & 28.4 & 30.5 & 45.0 & 41.5 & 51.0  \\
		
		SegMiF  & 85.9 & 66.3 & \textcolor{red}{\textbf{64.4}} & 34.0 & 34.7 & 43.0 &  \textcolor{red}{\textbf{48.0}} & 52.4  \\
		
		FILM & 86.6 & 66.4 & 61.9 & \textcolor{blue}{\textbf{36.3}}  & 32.0 & \textcolor{blue}{\textbf{46.7}} & \textcolor{blue}{\textbf{46.7}} & \textcolor{blue}{\textbf{52.7}}  \\
		
		SHIP & 86.7 & 63.9 & 59.7 & 33.5 & 35.2 & 44.4 & 42.4 & 51.5 \\
		
		MRFS & 86.1 & 61.5 & 59.7 & 30.6 & 35.2 & 44.9 & 38.4 & 50.4  \\
		
		EMMA & 86.1 & 65.6 & 63.8 & 25.4 & 29.1 & 44.9 & 41.3 & 51.7  \\
		
		TIMFusion & 86.4 & 58.7 & 60.3 & 29.8 & 33.6 & 44.1 & 38.3 & 49.8  \\
		
		SAGE  & {\textcolor{red}{\textbf{87.9}}} & \textcolor{red}{\textbf{67.5}} & \textcolor{blue}{\textbf{64.3}} & \textcolor{red}{\textbf{41.2}}  & 34.1 & \textcolor{red}{\textbf{46.8}} & \textcolor{blue}{\textbf{46.7}} & \textcolor{red}{\textbf{54.0}}  \\
		
		\bottomrule
	\end{tabular}
	\caption{Quantitative semantic segmentation results of SOTA approaches on the MFNet dataset.}
	\label{tab:mfnet}
	\vspace{-0.6cm}
\end{table}

\begin{figure*}[!htb]
	\centering
	\setlength{\tabcolsep}{1pt} 
	
	\includegraphics[width=0.99\textwidth, height=0.21\textheight]{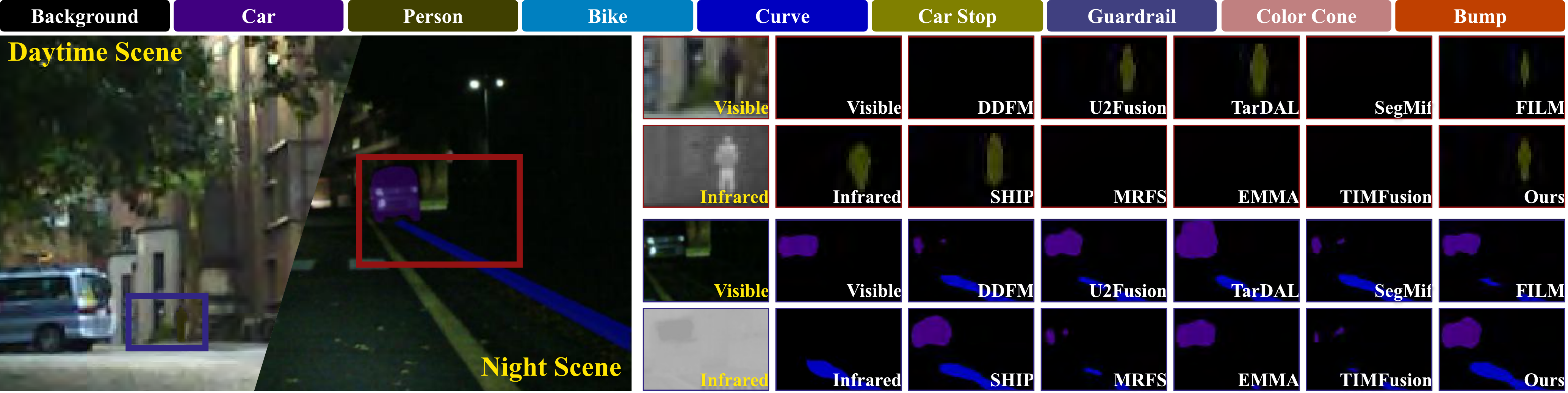}
	
	\caption{Qualitative segmentation results of SOTA approaches on the MFNet dataset.}
	\vspace{-0.15cm}
	\label{fig:seg_main_mfnet}
\end{figure*}

\subsection{Results of Multi-modality Segmentation}

\noindent \textbf{Qualitative Comparisons.}
We present the segmentation visualization results on the challenging new FMB dataset in Figure~\ref{fig:seg_main_fmb}. In the daytime intersection scene, thanks to the powerful semantic priors provided by SAM, our method is the only one to completely distinguish between the truck and the bus. In the nighttime road scene, we successfully segment the sidewalk, achieving optimal performance. Additionally, we showcase the segmentation results on the MFNet dataset in Figure~\ref{fig:seg_main_mfnet}. Leveraging the advanced semantic information learned by SAM, our method successfully segments distant small pedestrian targets during the daytime and near-invisible lane curve at night.

\noindent \textbf{Quantitative Comparisons.}
Table~\ref{tab:123} presents a qualitative comparison of IoU results for two segmentation frameworks on the FMB dataset. The first framework involves re-training images generated by various fusion methods, while the second uses the output from an open-vocabulary segmentation network with prompt words, without retraining. In traditional comparisons, our method achieves a 3.0 mIoU improvement over the second-best approach, demonstrating its competitive edge across categories. Additionally, in the segmentation network that requires no training, our method also performs favorably, benefiting from the adaptability provided by SAM's semantic information. Table~\ref{tab:mfnet} further showcases the qualitative results on the MFNet dataset, which has lower resolution and fewer labels. Although our method shows some gaps in background categories, it achieves overall superiority in key perception categories and mean performance. Our method integrates SAM with the fusion network, achieving a breakthrough in performance for efficient segmentation.

\begin{table}[h]
	\centering
	\small
	\renewcommand\arraystretch{0.9} 
	\setlength{\tabcolsep}{1.0mm}
		\begin{tabular}{|l|c|c|}
			\toprule
			\multirow{2}{*}{Variants} & \multicolumn{1}{c|}{mIoU} & \multicolumn{1}{c|}{Fusion Metrics} \\
			& \textbf{MFNet} / \textbf{FMB} & \textbf{EN} / \textbf{SD}/ \textbf{SCD} / \textbf{MS-SSIM} \\
			\bottomrule
			\multicolumn{3}{|c|}{I. The Impact of SAM} \\
			\toprule
			(a) w/o SAM& 50.7 / 56.5 &  \textcolor{blue}{\textbf{6.713}} / 36.93 / \textcolor{red}{\textbf{1.612}} / \textcolor{red}{\textbf{1.127}} \\
			(b) $\textbf{P}$ from $\mathbf{Y}_\text{label}$& 53.2 / 56.4 & 6.239 / 33.52 / 1.412 / 1.102 \\
			(c) Swap $\mathcal{S}^\text{O}$  &\textcolor{blue}{\textbf{53.4}} / 55.5 & 6.312 / 34.79 / 1.316 / 1.097
			\\
			\bottomrule
			\multicolumn{3}{|c|}{II. Study on the SPA Module} \\
			\toprule
			(a) \textit{w/o} \(\textbf{Z}\) & 52.3 / \textcolor{blue}{\textbf{57.2}} & 6.333 / 32.81 / 1.384 / 1.002 \\
			(b) \textit{w/o}  k-v in PR & 52.1 / 56.1 & 6.417 / 37.19 / 1.261 / 1.063 \\
			(c) \textit{w/o} PR & 53.3 / \textcolor{blue}{\textbf{57.2}} & 6.556 / 35.48 / 1.576 / 1.046 \\
			\bottomrule
			\multicolumn{3}{|c|}{III. Discussion on the Distillation Scheme} \\
			\toprule
			(a) w/o $\mathcal{L}_{\text{fea}}$ & 49.7 / 51.6 & 6.436 / 37.28 / 1.277 / 1.079 \\
			(b) w/o $\mathcal{L}_{\text{cont}}$ & 50.9 / 56.2 & 6.669 / 37.80 / 1.462 / 1.086 \\
			(c) w/o $\mathcal{L}_{\text{cs}}$ & 51.7 / 54.3 & 6.524 / \textcolor{blue}{\textbf{39.11}} / 1.356 / 1.038 \\
			(d) Offline Dist. & 50.0 / 50.4 & 6.221 / 32.65 / 1.581 / 0.987 \\
			\hline
			 SAGE & \textcolor{red}{\textbf{54.0}} / \textcolor{red}{\textbf{61.2}} & \textcolor{red}{\textbf{6.872}} / \textcolor{red}{\textbf{40.77}} / \textcolor{blue}{\textbf{1.604}} / \textcolor{blue}{\textbf{1.117}} \\
			\bottomrule
		\end{tabular}
	\caption{Quantitative results on both fusion and segmentation of all variants in three ablation studies.}
	\vspace{-0.3cm}
	\label{tab:abl}
\end{table}

\begin{table*}[ht]
	\centering
	\renewcommand\arraystretch{0.9} 
	\begin{tabular}{|l|cccccccccc|}
		\toprule
		{Method} & {DDFM} & {U2Fusion} & {TarDAL} & {SegMiF} & {FILM} & {SHIP} & {MRFS} & {EMMA} & {TIMFusion} & \textbf{Ours} 
		\\ 
		\bottomrule
		\toprule
		\textbf{Time(ms)} & 280K & 42.31 & \textcolor{blue}{\textbf{16.57}} & 150.5 & 183.3 & 27.93 & 95.86 & 25.73 & 17.39 & \textcolor{red}{\textbf{10.47}} 
		\\ 
		\textbf{FLOPs(G)} & 1.34M & 518.0 & 233.3 & 500.4 & 230.3 & 401.5 & 301.2 & 106.4 & \textcolor{blue}{\textbf{100.7}} & \textcolor{red}{\textbf{52.06}} 
		\\ 
		\textbf{Params(M)} & 552.7 & 0.659 & 0.297 & 0.621 & 0.491 & 0.525 & 133.4 & 1.516 & \textcolor{blue}{\textbf{0.158}} & \textcolor{red}{\textbf{0.136}} 
		\\ 
		\bottomrule
	\end{tabular}
	\caption{Efficiency comparison with other state-of-the-art methods on M$^3$FD benchmarks.}
	\label{tab:eff}
	\vspace{-0.4cm}
\end{table*}

\subsection{Ablation Studies}

\begin{figure}[!htb]
	\centering
	\setlength{\tabcolsep}{1pt} 
	
	\includegraphics[width=0.48\textwidth]{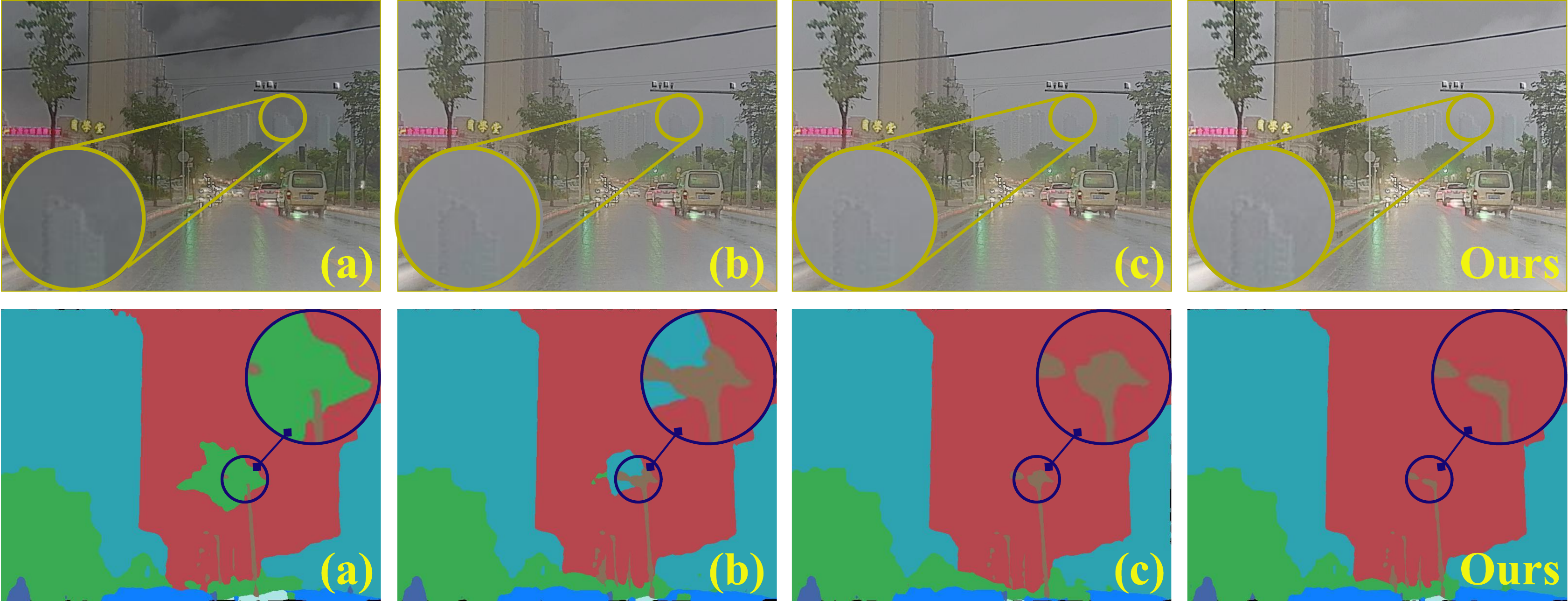}
	
	\caption{Visualization comparison of impact of SAM.}
	\label{fig:abl_sam}
	\vspace{-0.2cm}
\end{figure}

\noindent \textbf{The Impact of SAM.} 
The Segment Anything Module serves as a critical cornerstone in our method. To explore its impact, we derive three variants by isolating SAM from the core framework. Specifically, in variant (a), we replace the semantic patches with randomly cropped source image patches, thereby removing SAM from the main network $\mathcal{M}$. In variant (b), we assist the generation of semantic patches with segmentation labels. Additionally, we replace $\mathcal{S}_\text{O}$ with a conventional segmentation network~\cite{xie2021segformer}. Variant (a) enhances the source image information, leading to the optimal similarity metrics, SCD and MS-SSIM, as shown in Table~\ref{tab:abl}. Our method fully leverages the semantic priors provided by SAM, resulting in the best foggy building contours, as illustrated in the first row of Figure~\ref{fig:abl_sam}.

\begin{figure}[!htb]
	\centering
	\setlength{\tabcolsep}{1pt} 
	
	\includegraphics[width=0.48\textwidth]{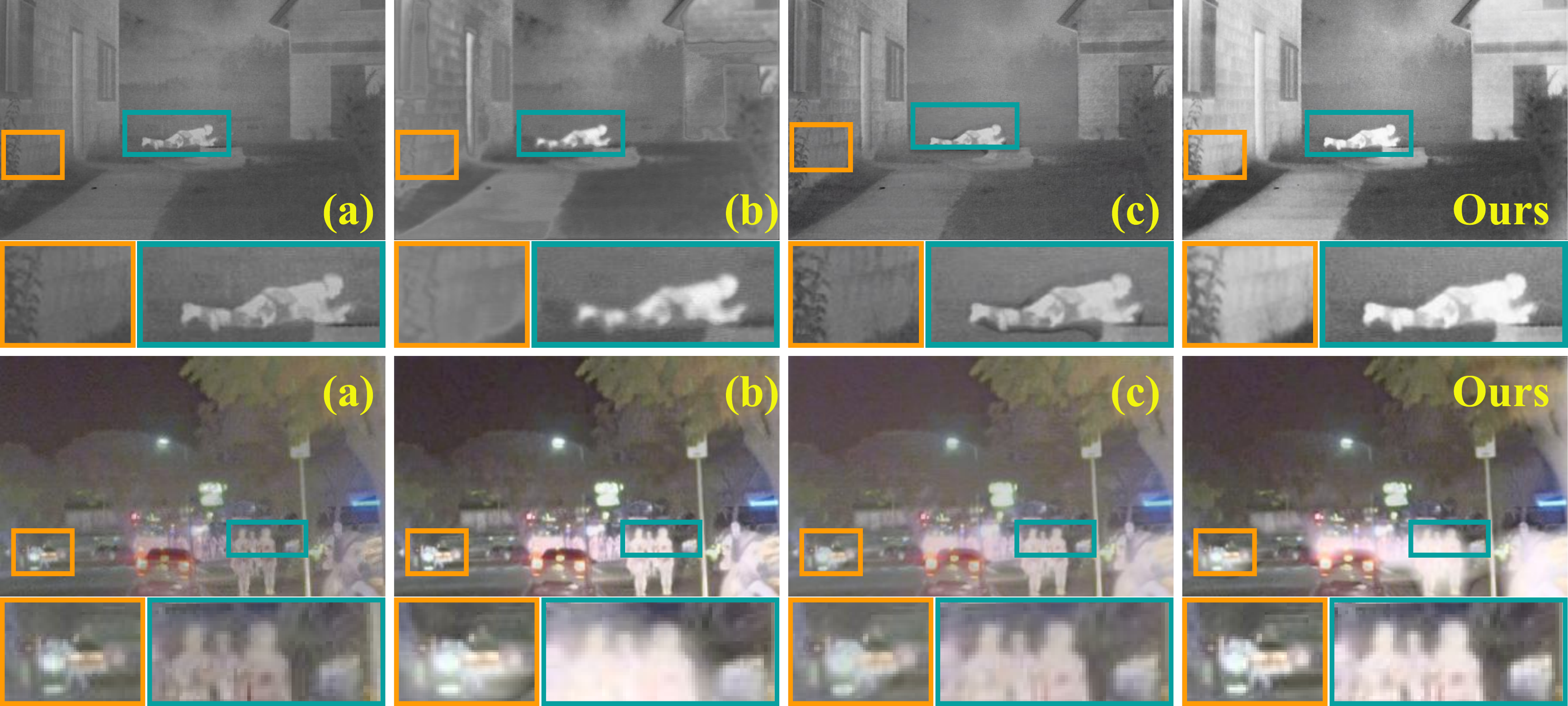}
	
	\caption{Visualization comparison of ablation on SPA.}
	\vspace{-0.2cm}
	\label{fig:abl_spa}
\end{figure}

\noindent \textbf{Study on the SPA Module.} 
The proposed Semantic Persistent Attention module plays a key role in integrating the semantic priors from SAM, effectively guiding the retention and enhancement of inherent modality features. To investigate its impact, we set up three variants: (a) \textit{w/o} latent representation \(\textbf{Z}\), (b) \textit{w/o} key-value pairs in the Persistent Repository, and (c) \textit{w/o} PR. The channel loss due to these modifications has been corrected. A visual comparison of these ablation variants is shown in Figure~\ref{fig:abl_spa}. It is evident that PR in the SPA module plays an essential role in maintaining critical information. The removal of any component leads to significant information loss. For instance, in (b), while the regions of interest are highlighted, the absence of source image information causes blurriness. Similarly, the absence of PR in the SPA module completely prevents the capture of beneficial semantic information, causing the network to lose focus on key areas and resulting in a low-contrast fusion results in (c).

\begin{figure}[!htb]
	\centering
	\setlength{\tabcolsep}{1pt} 
	\includegraphics[width=0.48\textwidth]{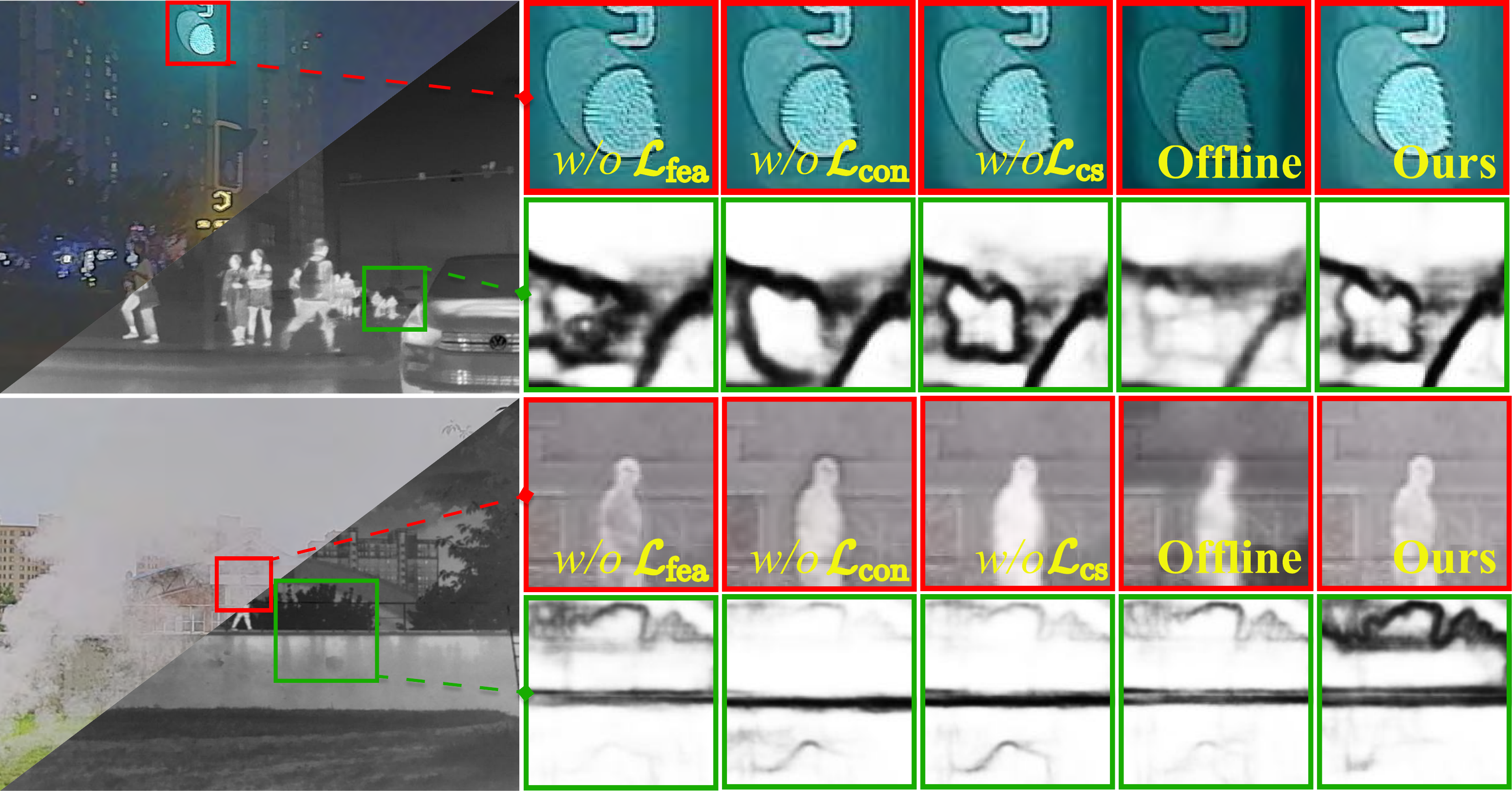}
	\caption{Visualization comparison of ablation on Distillation.}
	\label{fig:abl_dstill}
	\vspace{-0.4cm}
\end{figure}

\noindent \textbf{Discussion on the Distillation Scheme.}
Furthermore, we discuss the distillation scheme. First, we conduct ablation studies on each component of the triple loss function, leading to variants (a)-(c) as shown in Table~\ref{tab:abl} III. Additionally, we replace the bi-level optimization distillation method with the offline distillation approach, forming variant (d). A visual comparison of these variants is shown in Figure~\ref{fig:abl_dstill}. Notably, the results of the traditional online distillation are significantly inferior to our method, both in terms of visual quality and the gradient map generated by variant (b). This confirms the effectiveness of the proposed distillation method when incorporating semantic information.

\subsection{Computational Efficiency Analysis}
In the M$^3$FD benchmark, we compare our method with 9 other SOTA methods in terms of time, FLOPS, and parameters. As shown in Table~\ref{tab:eff}, our method demonstrates significant advantages across all aspects, particularly in time and FLOPs. Specifically, due to our distillation scheme, our sub-network is able to adjust flexibly while maintaining high efficiency, resulting in reduced FLOPs. Compared to other methods, our approach achieves a processing time of 10.47 ms and FLOPs of 52.06 G, outperforming most of the existing methods, with a parameter count of only 0.136M, showcasing computational efficiency.

Our method effectively alleviates the computational burden of SAM during inference, significantly reducing the computational overhead while preserving semantic information. This design allows our network to strike a balance between processing speed and computational resources, leading to superior efficiency in practical applications.

%% file: sec/5_Conc.tex
\section{Conclusion}

In this work, we propose a fusion method that utilizes semantic priors from the SAM for IVIF. We design the Semantic Persistent Attention module, which integrates semantic information while retaining source details. Additionally, we introduce a  bi-level distillation scheme with triplet loss to reduce computational complexity by decoupling  fusion from SAM during inference. Extensive experiments show that our method achieves superior fusion and task performance, outperforming across multiple datasets.

\section*{Acknowledgments}
\vspace{-0.2cm}
This work is partially supported by the National Natural Science Foundation of China (Nos. 62450072, U22B2052,\\ 62302078), Central Guidance for Local Science and Technology Development Fund (Youth Science Fund Project, Category A, No. 2025JH6/101100001), the Distinguished\\ Young Scholars Funds of Dalian (No. 2024RJ002), the Chi-\\na Postdoctoral Science Foundation (No. 2023M730741) and the Fundamental Research Funds for the Central Universities.\\
\vspace{-0.7cm}